\documentclass[journal]{IEEEtran}
\usepackage[T1]{fontenc}
\usepackage[utf8]{inputenc}
\usepackage{cite}
\usepackage{amsmath,amssymb,amsthm}
\usepackage{booktabs}
\usepackage{graphicx}
\usepackage{balance}
\usepackage{url}
\usepackage[hidelinks]{hyperref}
\graphicspath{{figures/}}

\newcommand{\RRTs}{\mbox{RRT$^{*}$}}
\newcommand{\Thetas}{\mbox{Theta$^{*}$}}
\newcommand{\Astar}{\mbox{A$^{*}$}}
\newcommand{\cstar}{c^{*}}
\newcommand{\Xfree}{X_{\mathrm{free}}}
\newcommand{\Jc}{J_{\eta,\tau}}
\newcommand{\Jk}{J^{*}_{\eta,\tau,k}}
\newcommand{\LS}{L_{S}}
\newtheorem{definition}{Definition}
\newtheorem{proposition}{Proposition}

\title{Does \RRTs{} Work on Uneven Terrain Meshes?}

\author{Otobong~Jerome and Tiago~Nascimento%
\thanks{The authors are with the Laborat\'orio de Engenharia de Sistemas e Rob\'otica, Universidade Federal da Para\'iba (UFPB), Jo\~ao Pessoa 58055-000, Para\'iba, Brazil (e-mail: o.jerome@laser.ci.ufpb.br; tiagopn@ci.ufpb.br).}%
\thanks{This work has been submitted to the IEEE for possible publication. Copyright may be transferred without notice, after which this version may no longer be accessible. Code: MeshNav3D 0.2.0 (release pending).}}

\begin{document}
\maketitle

\begin{abstract}
The optimal rapidly exploring random tree (\RRTs) is proven asymptotically optimal on open subsets of Euclidean space with straight-line local planning. We ask whether it works on a triangle-mesh terrain, a two-dimensional surface in $\mathbb{R}^3$. First, the original \RRTs{} cannot run on a mesh: the surface has zero volume, so uniform sampling never hits it, no clearance ball fits in it, and chords across a fold leave it. Second, the surface reformulation we call Mesh \RRTs{}, with its naive edge, the chord costed by its length, solved every run on heightfields with slopes up to $75^\circ$ but minimises the wrong objective: its cost, the straight-line length of a polyline whose vertices lie on the surface while its segments may leave it by a tolerance $\tau$, is the length of no route on the surface and lies strictly below the geodesic length on any curved mesh. In numeric simulation and after 30\,000 iterations the reported cost is on average up to 2.7\% below the exact geodesic, a cost no path attains, and the returned route on average 0.1--0.5\% longer than it. Rewiring keeps the biased edges, the bias decays only linearly with the step on polyhedral surfaces, and shrinking $\tau$ trades bias for connectivity. Third, replacing each chord by a path on the surface above it, built from projected points joined across the mesh edges they straddle and costed by its own length, makes every edge cost the exact length of a feasible path; with no other change the exact length of the returned routes is on average within 0.07--0.16\% of the geodesic at every roughness tested, two to 3.4 times closer than with chords on the rougher terrains. The planner is in MeshNav3D 0.2.0 (release pending).
\end{abstract}


\section{Introduction}
\IEEEPARstart{G}{round} robots on rough terrain and surface-bound aerial vehicles plan on a surface, usually a triangle mesh, and the natural cost is geodesic length. The established tools are graph planners on the mesh (Dijkstra, \Astar, \Thetas) and geodesic solvers such as the exact Mitchell--Mount--Papadimitriou (MMP) algorithm \cite{mmp1987}, fast marching \cite{kimmel1998} and the heat method \cite{crane2013}. Sampling-based optimal planners, led by the optimal rapidly exploring random tree (\RRTs) of Karaman and Frazzoli \cite{karaman2011}, are rarely applied here, and their guarantees are stated for Euclidean free space. This paper is about one of their primitives, \textsc{Line}, the routine that connects two tree nodes and prices the connection.

Jaillet and Porta extended \RRTs{} to implicitly defined configuration manifolds such as closed kinematic chains, with local charts built by higher-dimensional continuation, and argued asymptotic optimality on the manifold \cite{jaillet2012,jaillet2013}; here the manifold is an explicit triangle mesh whose faces are the charts and whose geodesics can be computed exactly. Mesh geodesics are a mature field \cite{mmp1987,surazhsky2005,crane2013}; exact MMP is our ground truth. Sampling-based mesh planning has been used for ground robots with the basic rapidly exploring random tree (RRT), where the geodesic computations between tree nodes are a heavy load that the method reduces \cite{witsmesh}; surface-bound aerial vehicles have instead been steered along the mesh by Riemannian motion policies on a planar parametrisation of it \cite{pantic2021}. \Astar{} on the mesh graph and \Thetas{} \cite{daniel2010} serve as graph-planner reference points. We know of no prior work measuring sampling-based mesh planners against exact geodesics or identifying the cost inconsistency studied here.

This paper keeps two quantities apart. The \emph{length} of a route is geometry: how far it runs along the mesh surface, measured here exactly with a geodesic solver. The \emph{cost} of a route is what the planner computes, optimises and reports: the number its \textsc{Line} primitive assigns to each edge, summed along the route. In Karaman and Frazzoli's setting the two coincide, because an edge is a straight segment inside the free space and its cost is its length. On a mesh they coincide only if the primitive makes them: a straight chord between two surface points is not a path on the surface, so its length is the length of no route, and a planner that charges it optimises and reports a cost below the surface length of the route it returns. Whether that happens on terrain meshes, by how much, and with what effect on the returned route is what we measure.

We test the hypothesis that \emph{\RRTs{} does not work on uneven terrain meshes} in its literal and in its practical form. Literally, \RRTs{} is the algorithm and guarantees of \cite{karaman2011}, defined on an open subset of $\mathbb{R}^d$, $d\ge 2$, with volume, uniform sampling and straight-line local planning. Practically, it is the reformulation an implementation on a mesh could make once the literal algorithm is found not to run there (Section~\ref{sec:why}): sample the surface, steer along it and, in the naive form, connect nodes by straight chords costed by Euclidean length.

Our contributions are:
\begin{enumerate}
  \item A structural argument that literal \RRTs{} cannot run on a mesh, because a surface has zero volume in $\mathbb{R}^3$ (Section~\ref{sec:why}).
  \item Mesh \RRTs{}, which follows Algorithm~6 of \cite{karaman2011} primitive by primitive, with two implementations of \textsc{Line}, naive (chords costed by their length) and consistent (surface paths costed by their own length), all else identical (Section~\ref{sec:algo}).
  \item The chord-polyline cost $\Jc$ (Definition~\ref{def:chord}) that the chord variant optimises, with proofs that it never exceeds the surface length of the same route (Proposition~\ref{prop:bias}) and that its optimum lies strictly below the geodesic length whenever the geodesic bends (Proposition~\ref{prop:gap}), and why rewiring keeps the biased edges (Section~\ref{sec:analysis}).
  \item Experiments against exact geodesics on terrains of controlled roughness: a dense-graph estimate of the chord optimum $\Jk$, the dependence of the bias on the step $\eta$ and the tolerance $\tau$, and the convergence of both variants (Sections~\ref{sec:setup}--\ref{sec:results}).
  \item An implementation in MeshNav3D 0.2.0 \cite{meshnav3d} (release pending).
\end{enumerate}

\section{Background}
\label{sec:background}

Karaman and Frazzoli \cite{karaman2011} take $X=(0,1)^d$, $d\ge 2$, an obstacle set $X_{\mathrm{obs}}$ and the free space $\Xfree=\mathrm{cl}(X\setminus X_{\mathrm{obs}})$, where $\mathrm{cl}$ denotes closure: a closed set with non-empty interior, hence positive Lebesgue measure $\mu(\Xfree)$. Paths are curves $\sigma:[0,1]\to\mathbb{R}^d$ of bounded variation; the cost $c(\sigma)$ is a monotone functional bounded by a multiple of the total variation, here Euclidean length. The goal $X_{\mathrm{goal}}$ is an open subset of $\Xfree$.

Algorithm~6 of \cite{karaman2011} grows a tree from $x_{\mathrm{init}}$ with five primitives per iteration: \textsc{SampleFree} draws uniformly in $\Xfree$; \textsc{Nearest} returns the closest tree vertex; \textsc{Steer} moves at most a step $\eta>0$ toward the sample; \textsc{Near} returns the vertices within radius
\begin{equation}
  r(n)=\min\left\{\gamma_{\mathrm{RRT}^{*}}\left(\frac{\log n}{n}\right)^{1/d},\ \eta\right\},
  \label{eq:radius}
\end{equation}
where $n$ is the number of tree vertices and $\gamma_{\mathrm{RRT}^{*}}$ a constant constrained by~\eqref{eq:gamma} below; and $\textsc{Line}(x_1,x_2)$ is the straight segment between points $x_1,x_2$, checked by \textsc{CollisionFree}. The new vertex $x_{\mathrm{new}}$ is attached to the vertex $x_{\mathrm{near}}$ among \textsc{Near} and the nearest vertex that has a collision-free connection and minimises $\mathrm{Cost}(x_{\mathrm{near}})+c(\textsc{Line}(x_{\mathrm{near}},x_{\mathrm{new}}))$, where $\mathrm{Cost}(v)$ is the cost of the tree path from $x_{\mathrm{init}}$ to a vertex $v$, and each vertex in \textsc{Near} is rewired through $x_{\mathrm{new}}$ when that lowers its cost.

Karaman and Frazzoli prove two guarantees under these definitions. Probabilistic completeness (Definition~14 of \cite{karaman2011}) needs a \emph{robustly feasible} problem: a solution path with strong $\delta$-clearance, i.e.\ every point $x$ of it has a closed ball $B(x,\delta)\subset\Xfree$ of radius $\delta$; otherwise the success probability is zero ``unless the samples are drawn from a singular distribution adapted to the problem''. Asymptotic optimality (Theorem~38) needs a robustly optimal solution and
\begin{equation}
  \gamma_{\mathrm{RRT}^{*}} > \big(2(1+1/d)\big)^{1/d}\left(\frac{\mu(\Xfree)}{\zeta_d}\right)^{1/d},
  \label{eq:gamma}
\end{equation}
$\zeta_d$ being the volume of the unit $d$-ball. Assumption~27 (the union of all optimal paths has measure zero) only excludes hitting the exact optimum in finitely many samples; it holds here, two points of a terrain mesh being joined by finitely many shortest paths, and plays no further role.

The proofs rely on Euclidean balls, on volume-uniform sampling, and on $c(\textsc{Line}(x_1,x_2))$ being the exact length of a feasible path in $\Xfree$: properties of the ambient space that the algorithm's control flow never uses.

\section{Why the Original \RRTs{} Cannot Run on a Mesh}
\label{sec:why}

Let $S\subset\mathbb{R}^3$ be a triangle-mesh terrain, $X=\mathbb{R}^3$ (or a bounding box) the ambient space and $\Xfree=S$. Each of the following then fails.
\begin{enumerate}
  \item \textbf{Zero measure.} $\mu(S)=0$ in $\mathbb{R}^3$, so uniform sampling in $X$ hits $S$ with probability zero. The completeness limit is zero, the case the authors exclude unless the sampler is ``adapted to the problem''.
  \item \textbf{No clearance.} No ball $B(x,\delta)$ fits inside a surface, so no problem on $S$ is robustly feasible; Theorems~23 and~38 of \cite{karaman2011} have no content.
  \item \textbf{Straight lines leave the surface.} For almost every pair of points on different non-coplanar faces, and every pair in the interiors of adjacent non-coplanar faces, $\textsc{Line}(x_1,x_2)\not\subset S$. With $\Xfree=S$, \textsc{SampleFree}, by definition the subsequence of uniform ambient samples falling in $\Xfree$ \cite[Sec.~3.1]{karaman2011}, is empty almost surely (item~1), so Algorithm~6 never produces a candidate vertex. Even with the ambient samples used directly, every steered point would be off $S$ almost surely (a sample within $\eta$ of the root is returned unchanged; a farther one lands on the sphere of radius $\eta$ about the root, which meets $S$ in a null set of directions) and \textsc{CollisionFree} would reject every edge.
\end{enumerate}
None of this depends on how uneven the terrain is. Any fold triggers item~3; only a perfectly flat mesh escapes, and on a flat mesh the algorithm is planar \RRTs, so the literal hypothesis holds for structural reasons.

An implementation on a mesh can instead reformulate the problem on the surface: $X=S$, a 2-manifold with geodesic metric $d_S$, area as the measure, $d=2$, $\zeta_2=\pi$ and $\mu(\Xfree)=\mathrm{area}(S)$, and the primitives have natural counterparts (Table~\ref{tab:primitives}). Area-uniform sampling is the singular distribution the paper allows for; clearance balls become geodesic disks, which exist on a terrain without holes, so robust feasibility returns. Only \textsc{Line}, the geodesic between two arbitrary surface points, lacks an exact cheap counterpart.

\begin{table*}[t]
\centering\footnotesize
\caption{Primitives of Algorithm~6 of \cite{karaman2011} and their mesh counterparts.}
\label{tab:primitives}
\begin{tabular}{@{}p{3.2cm}p{3.6cm}p{5.0cm}p{5.0cm}@{}}
\toprule
Primitive in \cite{karaman2011} & On the manifold $S$ & Chord implementation & Faithful? \\
\midrule
\textsc{SampleFree} (uniform in $\Xfree$) & uniform by area on $S$ & area-weighted face, uniform barycentric point & yes \\
\textsc{Nearest} (Euclidean) & geodesic nearest & Euclidean nearest & approximation; on a heightfield with slope $\le\psi$, $d_S(x_1,x_2)\le \|x_2-x_1\|/\cos\psi$ \\
\textsc{Steer} (move $\eta$ along a line) & move $\eta$ along the geodesic & chord step, projected back onto $S$, pulled back to $\le\eta$ & approximation; node lies on $S$ \\
\textsc{Near} (ball, radius $r(n)$) & geodesic disk, radius $r(n)$ & Euclidean ball $\cap\,S$, same radius law, $d=2$ & superset of the geodesic disk ($\|x_2-x_1\|\le d_S(x_1,x_2)$) \\
\textsc{Line} and its cost & geodesic segment, length $d_S$ & straight chord, length $\|x_2-x_1\|$ & \textbf{no}: not on $S$, and below $d_S$ \\
\textsc{CollisionFree} & segment inside $\Xfree$ & chord within a tolerance $\tau$ of $S$, optional slope limit & proxy for holes and forbidden faces \\
$X_{\mathrm{goal}}$ (open set) & a point, reached exactly & goal-biased sampling & standard practice \\
Parent choice, rewiring & Algorithm~6, lines 9--16 & identical, except that the nearest vertex is also a rewiring candidate and a coincident steered point re-chooses the existing vertex's parent instead of adding a duplicate; cost propagation to descendants & yes \\
\bottomrule
\end{tabular}
\end{table*}

Everything marked ``approximation'' in Table~\ref{tab:primitives} converges to the manifold primitive as the scale shrinks, since geodesic disks are Euclidean to first order. The \textsc{Line} row does not: the step $\eta$ does not shrink with $n$, so neither does the gap between a chord and the geodesic it stands for.

\section{Mesh \RRTs{} and the \textsc{Line} Primitive}
\label{sec:algo}

Mesh \RRTs{} follows Algorithm~6 of \cite{karaman2011}, changing how the primitives are realised on a triangle mesh and departing from the listing in three details: the nearest vertex is also a rewiring candidate; a new vertex is inserted whenever any candidate parent has a valid edge, not only when its edge from the nearest vertex is valid; and a steered point that coincides with an existing vertex (in practice the goal, at every goal-biased sample once the goal is in the tree) is not inserted again, that vertex's parent is re-chosen over its near set and the rewiring step is skipped for that sample. Closest-point queries use a static cell locator built once over the mesh.

\subsubsection*{Sampling} A face is drawn with probability proportional to its area and a point uniformly inside it by barycentric coordinates; with probability $p_{\mathrm{goal}}=0.1$ the goal itself is drawn.

\subsubsection*{Nearest and Near} Euclidean distance in $\mathbb{R}^3$, with radius $r(n)=\min\{\gamma_{\mathrm{RRT}^{*}}\sqrt{\log n/n},\,\eta\}$ and $\gamma_{\mathrm{RRT}^{*}}=1.1\times 2\sqrt{1.5}\times\sqrt{\mathrm{area}(S)/\pi}$, above the bound~\eqref{eq:gamma}. A Euclidean ball of radius $r$ contains the geodesic disk of radius $r$, so the near set is never too small.

\subsubsection*{Steer} Move at most $\eta$ from the nearest vertex toward the sample and project onto $S$. Since projection can stretch the chord, the point is pulled back along it and re-projected (up to six times); a point still beyond $\eta$ is discarded, so every tree edge is at most one step long.

\subsubsection*{Edge validity} Points along the chord between two nodes must lie within a tolerance $\tau$ of $S$ ($\tau=0.15\,\eta$ unless stated); this rejects chords that tunnel through bumps or bridge gaps. The chord implementation tests $k=3$ uniformly spaced interior points per step length; the surface implementation tests every point at which it bisects the chord (below, and Section~\ref{sec:ksweep}). An optional slope limit rejects edges steeper than a given angle.

\subsubsection*{Goal} The goal joins the tree only when a steered point lands on it exactly, which goal-biased sampling brings about once any node lies within $\eta$ of it. There is no straight connection from a distance.

\subsubsection*{Parent choice and rewiring} Lines 9--16 of Algorithm~6 of \cite{karaman2011}, with the nearest vertex added to the near set for rewiring as well as for the parent choice. Candidates are scanned in order of $\mathrm{Cost}(x_{\mathrm{near}})$ plus chord length, stopping once that lower bound cannot beat the best valid edge found. A rewire updates the costs of all descendants.

\subsubsection*{Step} $\eta$ defaults to the mean mesh edge length, 0.57 on the reference terrain.

\subsection{Two implementations of \textsc{Line}}
\RRTs{} is a single planner and \textsc{Line} one routine inside it. Our two implementations differ only in what an edge \emph{is} and what it \emph{costs}, so that any difference in behaviour is attributable to this primitive alone.
\begin{itemize}
  \item \textbf{Chord (naive).} The straight segment between two surface points, costed by its Euclidean length: the obvious implementation, although the segment itself is not on the surface.
  \item \textbf{Surface (consistent).} The edge is built from the chord's endpoints by bisection: the chord's midpoint is projected onto $S$ and the halves are treated recursively until consecutive points lie on one face or on two faces sharing an edge. Points on one face are joined by their segment; points on edge-adjacent faces by the shortest path across the shared edge, the straight line in the unfolding of the two faces, with the crossing point inserted (or, if that line misses the edge, through its nearer endpoint); points on faces sharing only a vertex are bisected further and joined through the vertex when it lies on the segment between them or when a cap of six bisections is reached. A projected point farther than $\tau$ from $S$ rejects the edge, as does a pair on faces sharing no vertex after six bisections, where the projection jumps because the chord passes beneath a ridge. Every vertex of the resulting polyline lies on $S$ and consecutive vertices share a face, so it is a path on $S$, and its length, the edge cost, is its exact surface length. The returned path concatenates these polylines, so the reported cost is the exact length of the path returned.
\end{itemize}
The surface variant costs 1.2 times the chord's time per iteration on the reference terrain and 1.4 and 1.5 times at amplitudes 2 and 4 (Table~\ref{tab:k}).

\section{Analysis: What Each Variant Optimises}
\label{sec:analysis}

\subsection{The chord-polyline cost}

Definition~\ref{def:chord} makes the two quantities of the Introduction precise for a route of the chord variant: $\Jc$ is its cost, the number the planner optimises, and $\LS$ is its length, the shortest distance a robot could travel on the surface through the same nodes.

\begin{definition}[Admissible chord polyline and its two lengths]
\label{def:chord}
Let $S$ be a finite, connected triangle mesh (a compact connected polyhedral surface), $\eta>0$ a step and $\tau\ge 0$ a tolerance. A \emph{chord polyline} from $a$ to $b$ is a finite sequence $P=(x_0,x_1,\dots,x_m)$ of points of $S$ with $x_0=a$ and $x_m=b$ such that each segment $[x_i,x_{i+1}]$ has length at most $\eta$ and lies within distance $\tau$ of $S$ at every point. Its \emph{chord cost} is
\begin{equation}
  \Jc(P)=\sum_{i=0}^{m-1}\|x_{i+1}-x_i\|,
  \label{eq:J}
\end{equation}
and its \emph{surface length} is
\begin{equation}
  \LS(P)=\sum_{i=0}^{m-1} d_S(x_i,x_{i+1}),
  \label{eq:LS}
\end{equation}
the length of the shortest route on $S$ that visits the nodes in order. $\Jc^{*}(a,b)$ is the infimum of $\Jc$ over admissible chord polylines and $\cstar=d_S(a,b)$ the geodesic distance.
\end{definition}

The chord variant of Section~\ref{sec:algo} grows a tree whose edges are these segments at their Euclidean lengths, with the distance condition checked at the $k$ interior points of Section~\ref{sec:algo} rather than at every point (Section~\ref{sec:ksweep} measures the effect of $k$), so its reported path cost is $\Jc$ of the returned polyline; the shortest a robot could travel on $S$ through the same nodes is $\LS$, the quantity measured in Section~\ref{sec:results}.

\begin{proposition}[Chord cost versus surface length]
\label{prop:bias}
For every admissible chord polyline $P$, $\Jc(P)\le\LS(P)$, with equality if and only if every segment $[x_i,x_{i+1}]$ lies on $S$. Moreover $\LS(P)\ge\cstar$.
\end{proposition}
\begin{IEEEproof}
On a compact connected surface a shortest path between any two points exists. The shortest path on $S$ from $x_i$ to $x_{i+1}$ is a rectifiable curve in $\mathbb{R}^3$ with those endpoints, so its length $d_S(x_i,x_{i+1})$ is at least $\|x_{i+1}-x_i\|$, with equality only if that path is the straight segment, the unique curve of minimal length between two points of $\mathbb{R}^3$, which then lies on $S$. Conversely, a segment lying on $S$ is a path on $S$ of length equal to the chord. Summing over $i$ gives the first claim; the triangle inequality for $d_S$ gives $\LS(P)\ge d_S(a,b)=\cstar$.
\end{IEEEproof}

\begin{proposition}[The chord optimum lies below the geodesic]
\label{prop:gap}
Let $S$ be a finite, connected triangle mesh, $\eta>0$, $\tau>0$, and let $\sigma^{*}$ be a shortest path on $S$ from $a$ to $b$, parametrised by arc length $s$. If $\sigma^{*}$ is not a single straight segment, then $\Jc^{*}(a,b)<\cstar$. If $\sigma^{*}$ is a straight segment, then $\Jc^{*}(a,b)=\cstar$. (The dichotomy does not depend on the choice of $\sigma^{*}$: if one shortest path is straight then $\cstar=\|b-a\|$ and every shortest path is that segment, so the first case is exactly $[a,b]\not\subset S$.) In either case, every value $\Jc(P)<\cstar$ is a cost that no path on $S$ from $a$ to $b$ attains.
\end{proposition}
\begin{IEEEproof}
$\sigma^{*}$ is polygonal, and its tangent direction can change only where it crosses a mesh edge transversally with non-zero dihedral angle or passes through a vertex \cite{mmp1987}; call a point where the direction does change a \emph{bend}. A shortest path that is not a single straight segment has at least one bend. Suppose $\sigma^{*}$ bends at arc length $s_0$, and let $\varepsilon>0$ be smaller than $\min\{\eta/2,\tau\}$ and than the two straight pieces adjacent to the bend. Form $P$ from the points of $\sigma^{*}$ at every bend, at arc lengths $s_0-\varepsilon$ and $s_0+\varepsilon$ in place of $s_0$, and at further points along the straight pieces so that consecutive nodes are at most $\eta$ apart. Every segment of $P$ other than $[\sigma^{*}(s_0-\varepsilon),\sigma^{*}(s_0+\varepsilon)]$ lies on $\sigma^{*}\subset S$ and is admissible. The exceptional segment has length strictly less than $2\varepsilon\le\eta$, because $\sigma^{*}$ bends at $s_0$, and every point of it lies within $\varepsilon\le\tau$ of one of its endpoints, which are on $S$; so $P$ is admissible, and
\[
  \Jc(P)=\cstar-\big(2\varepsilon-\|\sigma^{*}(s_0+\varepsilon)-\sigma^{*}(s_0-\varepsilon)\|\big)<\cstar .
\]
Hence $\Jc^{*}\le\Jc(P)<\cstar$. If $\sigma^{*}$ is a straight segment, then $\cstar=\|b-a\|\le\Jc(P)$ for every polyline $P$ by the triangle inequality in $\mathbb{R}^3$, and subdividing $[a,b]\subset S$ into segments of length at most $\eta$ gives an admissible polyline with $\Jc=\|b-a\|$; so $\Jc^{*}=\cstar$. Finally, every path on $S$ from $a$ to $b$ has length at least $\cstar$, so no value below $\cstar$ is attained.
\end{IEEEproof}

Unless $[a,b]$ itself lies on $S$, the first case of Proposition~\ref{prop:gap} applies; on a terrain without large flat regions this is typical for points more than a face diameter apart, and it holds for all 20 scenario pairs on the curved terrains of Section~\ref{sec:setup}, whose geodesics exceed the straight-line distance by 2--29\%. The chord variant thus optimises a quantity whose optimum is strictly below the length of every feasible path; with its sampled distance test it optimises $\Jc$ over a superset of the admissible polylines, whose optimum $\Jk$ satisfies $\Jk\le\Jc^{*}<\cstar$ and is the quantity Section~\ref{sec:results} estimates. The gap $\cstar-\Jc^{*}$ is a property of the cost, independent of the sampler, and Proposition~\ref{prop:gap} does not quantify it: how far below $\cstar$ the optimum lies depends on $\eta$ and $\tau$ (Section~\ref{sec:scaling}) and is estimated through $\Jk$ in Section~\ref{sec:results}. With $\tau=0$ only segments on $S$ are admissible, so $\Jc=\LS$, but nodes on adjacent non-coplanar faces can then be joined only through a node exactly on their common edge, which area-uniform sampling never produces: in the dense-graph experiment of Section~\ref{sec:results} with $\tau\to 0$ only the 9--12\% of candidate pairs within a single face are admissible and start and goal are disconnected in all ten rough-terrain scenarios.

\subsection{Why rewiring does not remove the bias}

The gap does not vanish as $n$ grows, although $r(n)$ shrinks and short chords approximate geodesics well. Rewiring (line~15 of Algorithm~6 in \cite{karaman2011}) replaces an edge only when a strictly cheaper alternative appears, and under $\Jc$ a single chord of length up to $\eta$ that cuts below a bend is cheaper than any chain of short edges following the surface, because it omits the length it skips; the cutting chords are therefore the edges rewiring keeps. Any asymptotic-optimality guarantee that transfers to the surface does so for $\Jc$, the only cost the tree evaluates: the reported cost would converge to $\Jk<\cstar$ and the routes to $\Jc$-minimising polylines, whose length $\LS$ exceeds $\cstar$ unless they are geodesics. We do not prove such a transfer (adapting Theorem~38 to the surface is open); Section~\ref{sec:results} estimates $\Jk$ independently of the tree and shows the tree's cost approaching it.

\subsection{Consistency and the surface primitive}

A \textsc{Line} primitive is \emph{consistent} if the cost it charges for an edge is the length of a feasible path between its endpoints. The straight line in \cite{karaman2011} is consistent by construction; the chord is not (Proposition~\ref{prop:bias}). The surface primitive is consistent exactly: every vertex of its polyline lies on $S$ and consecutive vertices share a face, so the polyline is a path on $S$, its length is that path's length, and the cost it charges can never fall below the surface distance between the edge's endpoints. The surface path through the projected points is not the geodesic between the endpoints; it is longer, by 0.002, 0.01 and 0.05\% of $\cstar$ on average on the routes returned at amplitudes 1, 2 and 4, and adding uniformly spaced samples along the chord lengthens it further, because a denser polyline follows the projection of the chord more closely. The planner brings the route to the geodesic by where it places nodes, so the primitive needs no sample count; bisection from the endpoints is enough (Section~\ref{sec:ksweep}). Where the projection of a chord jumps between faces sharing no vertex, there is no surface path along that chord and the primitive rejects the edge; on the terrains tested this never happened. Every rejection was a tolerance failure at a bisection point, 1.2\% of candidate edges at amplitude 4 and 0.1\% at amplitude 2 against 1.4\% and 0.1\% for the chord's tolerance test, and on average the first solution came no later than the chord's (131 against 140 iterations at amplitude 4; Section~\ref{sec:results}).

\subsection{Scaling with the step and with the tolerance}
\label{sec:scaling}

\subsubsection*{Smooth surfaces} For a geodesic of roughly constant curvature $\kappa$ sampled at arc spacing $\eta$, each chord falls short of its arc by $\kappa^2\eta^3/24+O(\kappa^4\eta^5)$, so the relative gap of the whole path is $\kappa^2\eta^2/24$, provided the sagitta $\kappa\eta^2/8$ stays within $\tau$.

\subsubsection*{Polyhedral surfaces} A mesh geodesic is a polyline that bends at mesh edges. A chord spanning a bend of turning angle $\theta$ at the middle of a step of length $\eta$ is shorter than the two arms by $\eta(1-\cos(\theta/2))=\eta\theta^2/8+O(\eta\theta^4)$; a bend splitting the step into arms $\alpha+\beta=\eta$ gives $\alpha\beta\,\theta^2/(2\eta)$, between $0$ (bend at a node) and $\eta\theta^2/8$ (centred), so $\eta\theta^2/8$ per bend is an upper envelope over sampling phase. Summed over the bends a route crosses, whose number is fixed by the mesh, the gap is \emph{linear} in $\eta$ as long as each step straddles at most one bend and the tolerance does not bind. The tolerance binds when the chord's deepest point exceeds $\tau$; for a centred crossing that point is at most $(\eta/4)\sin\theta$ from $S$, with equality when the path crosses the edge perpendicularly, so it can bind only for $\eta>4\tau/\sin\theta$, from that value on for perpendicular crossings and later for oblique ones; once it binds, the per-bend gap saturates near $\tau\theta/(2\cos(\phi/2))$, where $\phi$ is the dihedral turning angle of the fold, which equals $\theta$ for a perpendicular crossing and exceeds it for an oblique one. With $\tau$ proportional to $\eta$, as in our experiments, the cap is itself proportional to $\eta$ and the linear law persists. Halving the step therefore only halves the gap on a mesh, while \RRTs{} in two dimensions needs roughly four times as many samples to cover the surface at half the spacing. Section~\ref{sec:step} measures the fixed-geodesic shortfall and finds a ratio of 1.5--2.3 per doubling of $\eta$.

\subsubsection*{Tolerance} A chord of length $\ell\le\eta$ that straddles a fold of turning angle $\theta$ symmetrically has its deepest point, on the bisector, at most $(\ell/2)\sin(\theta/2)$ from $S$ (equivalently $(\eta/4)\sin\theta$ when $\eta$ is the surface length of its two arms), with equality for a crossing perpendicular to the fold edge and less for oblique crossings, so it is admissible whenever $\ell\le 2\tau/\sin(\theta/2)\approx 4\tau/\theta$. Asymmetric chords with one endpoint within $\tau/\sin\theta$ of the fold are admissible at any length up to $\eta$, but the per-fold shortfall of any admissible chord is at most about $\tau\theta/(2\cos(\phi/2))$, and the combined per-fold gap about $\min\{\eta,4\tau/(\theta\cos(\phi/2))\}\,\theta^2/8$; both reduce to $\tau\theta/2$ and $\min\{\eta,4\tau/\theta\}\,\theta^2/8$ for perpendicular crossings. Shrinking $\tau$ therefore shrinks the gap, but it removes the symmetric crossings that carry connectivity at a given sample density faster than it removes the bias, so at any finite sample density it trades one against the other (Section~\ref{sec:tol}). The surface primitive has no such trade-off, because $\tau$ only guards the projection and the cost is the polyline's length for every admissible edge.

\subsection{The flat case}
On the flat control, a grid split along one diagonal so that every two edge-adjacent faces form a convex quadrilateral, every chord lies on $S$ and $\Jc=\LS$; the two implementations charge the same cost for every edge except where the surface primitive's bisection cap routes through a vertex, both are planar \RRTs, and both converge to the same cost (Section~\ref{sec:results}), so the discrepancy between them on curved terrain is caused by curvature.

\section{Experimental Setup}
\label{sec:setup}

\subsubsection*{Terrains} Synthetic egg-crate heightfields $z=A\sin(\omega x)\sin(\omega y)$ on a $10\times 10$ square, triangulated on a $23\times 23$ grid (529 vertices, 968 faces), with wavenumber $\omega=2\pi\times 1.5/10$, i.e.\ 1.5 periods per side; the amplitude $A$ sets the roughness. $A=1$ matches the grid, extent and height range of the terrain bundled with MeshNav3D, $z=\sin x\cos y$, which has 1.59 periods per side and a $45^\circ$ maximum slope against 1.5 and $43^\circ$ here (mean edge 0.57 in both); $A=4$ has mean edge 1.01 and slopes up to $75^\circ$; $A=0$ is a flat plane and is the control.

\subsubsection*{Ground truth} The exact geodesic $\cstar$ between two mesh vertices from the MMP algorithm \cite{mmp1987,surazhsky2005} (pygeodesic), the engine of MeshNav3D's MMP planner.

\subsubsection*{Scenarios} Five random (start, goal) vertex pairs at least half the diagonal apart in the plane, drawn once and reused at every amplitude so that only roughness changes between terrains.

\subsubsection*{Planners} Mesh \RRTs{} with the chord and with the surface implementation of \textsc{Line}, step $\eta$ equal to the mean edge length, $\tau=0.15\,\eta$, three seeds per scenario, 30\,000 iterations per run; the best cost after every iteration is recorded, so one run yields the whole convergence curve. Graph references: \Astar{} on the mesh edge graph and \Thetas{} \cite{daniel2010}, both from MeshNav3D.

\subsubsection*{Metrics} Success; the planner's \emph{reported} cost over $\cstar$; the \emph{surface length} $\LS$ of the returned route over $\cstar$; the iteration of the first solution; and the fraction of candidate edges rejected by the tolerance test. For a chord route $\LS$ is computed exactly as in~\eqref{eq:LS}: the route's nodes are inserted as mesh vertices by splitting the triangles or edges they lie on, which leaves the surface unchanged, and MMP geodesics are summed between consecutive nodes. A surface route is measured the same way through its tree nodes; its reported cost, the polyline length, is the exact length of the returned path itself and therefore at least the geodesic length through the same nodes. For \Thetas{} routes, whose nodes are mesh vertices, $\LS$ sums vertex-to-vertex MMP geodesics; \Astar{} routes run along mesh edges and are exact as returned.

\subsubsection*{Limit experiment} To estimate $\Jk$ without the randomness of the tree, we sample $N$ points uniformly on the surface, connect every pair closer than $\eta$ that passes the edge test, weight each edge by the implementation's cost, and run Dijkstra between the scenario vertices; as $N\to\infty$ this approaches the optimum of that cost over edges passing the same test, for either implementation.

\subsubsection*{Step-scaling experiment} The exact geodesic is resampled at spacing $\eta$ and its chord-polyline length compared with $\cstar$. This isolates the chord gap for a fixed path.

\section{Results}
\label{sec:results}

\begin{figure*}[t]
\centering
\includegraphics[width=\textwidth]{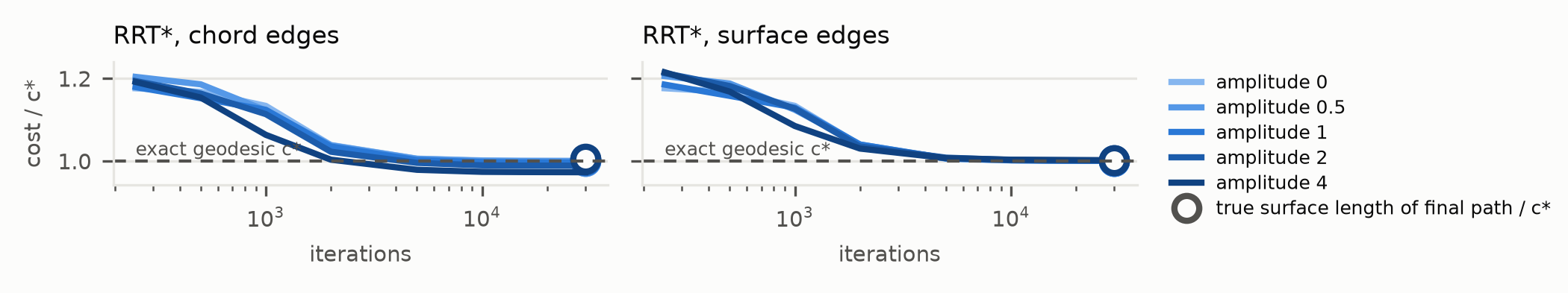}
\caption{Mean reported cost over the exact geodesic $\cstar$ against iterations (log scale), 5 scenarios $\times$ 3 seeds per terrain; chord edges left, surface edges right. Rings: exact surface length $\LS$ of the final route. Below the dashed line the reported cost is shorter than any path on the surface.}
\label{fig:conv}
\end{figure*}

Chord \RRTs{} found a solution in every run and converges, but to a reported cost below the geodesic optimum; the surface variant converges to the optimum. Fig.~\ref{fig:conv} shows the mean reported cost against iterations for both \textsc{Line} implementations at five roughness levels.

On the flat control both variants converge to the same cost, 1.0007 of the straight line, the planar behaviour proven in \cite{karaman2011}. From amplitude 0.5 upward the chord curves cross below $\cstar$ and keep falling as the tree densifies, to 0.996, 0.989 and 0.973 at amplitudes 1, 2 and 4, while the rings, the mean exact surface length of the returned routes, are 1.0008, 1.0014, 1.0033 and 1.0055 of $\cstar$ at amplitudes 0.5, 1, 2 and 4: on average 0.1--0.5\% longer than the geodesic; the excess grows with roughness (individual chord routes are 0.04--1.1\% long and their reported costs reach 3.3\% below $\cstar$; individual surface routes are 0.03--0.5\% long). The surface curves settle just above $\cstar$; their final routes measure 1.0007, 1.0007, 1.0011 and 1.0016 of $\cstar$, two to 3.4 times closer to the geodesic than the chord routes at amplitudes 1--4 (1.2 times at amplitude 0.5, where both are within 0.1\%), and their reported costs (1.0007, 1.0007, 1.0012 and 1.0021) lie at or above those exact lengths, as the length of a feasible path must. Two scenarios per rough terrain run to 60\,000 iterations change the chord cost by at most $2\times10^{-4}$ of $\cstar$ between 30\,000 and 60\,000 iterations (a plateau at $\Jk$), while the surface cost still falls by up to $2.7\times10^{-3}$.

\subsection{What each implementation converges to}

The dense-graph estimate of each optimum ($N=8000$, Table~\ref{tab:limit}) confirms that each tree converges to the optimum of its own cost. The chord optimum $\Jk$ lies below $\cstar$, as Proposition~\ref{prop:gap} requires, and moves further below as the graph densifies: at amplitude 2 from 1.015 at $N=2000$ to 0.986 at $N=16\,000$, while the exact surface length of the minimiser falls from 1.025 to 1.003 of $\cstar$ as sparse-graph detours disappear, so the gap between what the cost reports and what the route measures widens with density. The surface optimum approaches $\cstar$ from above (1.025 to 1.001 over the same range); its minimiser's exact length is within 0.24\% of $\cstar$ at every amplitude for $N=8000$, and the optimum lies at or above it (1.0020 against 1.0019, 1.0026 against 1.0024 and 1.0029 against 1.0019 at amplitudes 1, 2 and 4), as it must for a cost that is the length of a feasible path.

\begin{table}[t]
\centering\footnotesize
\caption{Dense-graph optimum of each edge cost ($N=8000$, $\tau=0.15\eta$) and the exact surface length $\LS$ of the route attaining it, over $\cstar$. Shorter of the two in bold.}
\label{tab:limit}
\setlength{\tabcolsep}{4pt}
\begin{tabular}{@{}lcccc@{}}
\toprule
Ampl. & Chord $\Jk/\cstar$ & Chord $\LS/\cstar$ & Surf. opt.$/\cstar$ & Surf. $\LS/\cstar$ \\
\midrule
1 & 0.9975 & 1.0025 & 1.0020 & \textbf{1.0019} \\
2 & 0.9890 & 1.0043 & 1.0026 & \textbf{1.0024} \\
4 & 0.9733 & 1.0044 & 1.0029 & \textbf{1.0019} \\
\bottomrule
\end{tabular}
\end{table}

\subsection{The gap is linear in the step}
\label{sec:step}

Resampling the exact geodesic at spacing $\eta$ and summing the chords gives the shortfall in Table~\ref{tab:step}, as a fraction of $\cstar$. The ratio per doubling of $\eta$ is about 2, whereas the smooth-surface expansion of Section~\ref{sec:scaling} predicts 4; on a mesh, halving the step halves the gap at roughly four times the sampling cost.

\begin{table}[t]
\centering\footnotesize
\caption{Chord shortfall of the exact geodesic resampled at spacing $\eta$, over $\cstar$ (mean of five scenarios).}
\label{tab:step}
\begin{tabular}{@{}lccccc@{}}
\toprule
Ampl. & $\eta=0.125$ & $0.25$ & $0.5$ & $1.0$ & Ratio per doubling \\
\midrule
0.5 & 0.0003 & 0.0007 & 0.0015 & 0.0034 & 2.1--2.3 \\
1   & 0.0009 & 0.0018 & 0.0039 & 0.0091 & 1.9--2.3 \\
2   & 0.0020 & 0.0042 & 0.0068 & 0.0145 & 1.6--2.1 \\
4   & 0.0043 & 0.0077 & 0.0116 & 0.0211 & 1.5--1.8 \\
\bottomrule
\end{tabular}
\end{table}

\subsection{Shrinking the tolerance trades bias for connectivity}
\label{sec:tol}

Table~\ref{tab:tol} repeats the dense-graph estimate for the chord cost at four tolerances. Tightening $\tau$ from $0.15\eta$ to $0.05\eta$ moves $\Jk$ toward $\cstar$ but leaves it below (0.997 and 0.990 at amplitudes 2 and 4). At $0.02\eta$ the estimate rises above $\cstar$, but the cost has not become consistent: 47--52\% of the candidate edges are no longer admissible, the sparser graph detours, and the selected route's surface length worsens to 1.0--1.6\%. At $0.005\eta$ three quarters of the edges are gone and routes are 14--23\% long. Doubling the sample density to $N=16\,000$ restores connectivity at $0.02\eta$ and the estimate drops back below $\cstar$ (0.999 and 0.995 at amplitudes 2 and 4): tightening $\tau$ only raises the sample density at which the bias appears. No tolerance gives both a consistent cost and a well-connected graph, as Section~\ref{sec:scaling} predicts.

The planner behaves the same way (Table~\ref{tab:tolrrt}). Tightening $\tau$ from $0.15\eta$ to $0.02\eta$ raises the share of rejected candidate edges from 0.1--1.4\% to 58--63\%, delays the first solution 28- to 61-fold (from 71--140 to 3900--4300 iterations on average), leaves only 1--2 of 15 runs solved after 2000 iterations, and still leaves the mean reported cost at or below $\cstar$ after 30\,000 iterations, while the exact length of the returned routes stays in the same 0.2--0.6\% band as at $\tau=0.15\eta$.

\begin{table}[t]
\centering\footnotesize
\caption{Chord \RRTs{} at three tolerances (5 scenarios $\times$ 3 seeds): rejected candidate edges, runs solved after 500 / 2000 / 5000 iterations, first-solution iteration (mean / worst), reported cost and exact surface length of the final route over $\cstar$.}
\label{tab:tolrrt}
\scriptsize
\setlength{\tabcolsep}{3pt}
\begin{tabular}{@{}llccccc@{}}
\toprule
Ampl. & $\tau/\eta$ & Rej. & Solved 500/2000/5000 & First sol. & Reported$/\cstar$ & $\LS/\cstar$ \\
\midrule
2 & 0.15 & 0.1\% & 15/15/15 & 71/120    & 0.9888 & 1.0033 \\
2 & 0.05 & 21\%  & 5/14/15  & 734/2114  & 0.9938 & 1.0023 \\
2 & 0.02 & 58\%  & 0/1/10   & 4318/8304 & 0.9999 & 1.0032 \\
\addlinespace
4 & 0.15 & 1.4\% & 15/15/15 & 140/385   & 0.9729 & 1.0055 \\
4 & 0.05 & 32\%  & 1/15/15  & 1168/1948 & 0.9870 & 1.0042 \\
4 & 0.02 & 63\%  & 0/2/11   & 3889/6411 & 0.9963 & 1.0060 \\
\bottomrule
\end{tabular}
\end{table}

\begin{table}[t]
\centering\footnotesize
\caption{Chord edges on the dense graph ($N=8000$) at four tolerances: admissible edges, estimated $\Jk$ and exact surface length of its minimiser, over $\cstar$ (mean of five scenarios).}
\label{tab:tol}
\begin{tabular}{@{}llrcc@{}}
\toprule
Ampl. & $\tau/\eta$ & Edges & $\Jk/\cstar$ & $\LS/\cstar$ \\
\midrule
2 & 0.15  & 288\,136 & 0.9890 & 1.0043 \\
2 & 0.05  & 241\,075 & 0.9966 & 1.0039 \\
2 & 0.02  & 154\,320 & 1.0066 & 1.0101 \\
2 & 0.005 &  72\,814 & 1.1429 & 1.1441 \\
\addlinespace
4 & 0.15  & 374\,856 & 0.9733 & 1.0044 \\
4 & 0.05  & 284\,247 & 0.9898 & 1.0048 \\
4 & 0.02  & 182\,789 & 1.0069 & 1.0159 \\
4 & 0.005 &  82\,446 & 1.2264 & 1.2290 \\
\bottomrule
\end{tabular}
\end{table}

\subsection{Success is unaffected by roughness}

Table~\ref{tab:complete} reports success and time to first solution for the chord implementation. The surface implementation also solves every run, with the same mean first-solution iteration as the chord up to amplitude 2 and an earlier one at amplitude 4 (131 against 140, 385 at worst for both); its tolerance test at bisection points rejects 1.2\% of candidate edges there against 1.4\% for the chord's, and no edge was rejected for a projection jump. Roughness doubles the time to a first solution at $75^\circ$ slopes but never prevents one, and at $\tau=0.15\eta$ the tolerance test rejects on average at most 1.4\% of candidate edges (2.0\% in the worst single run).

\begin{table}[t]
\centering\footnotesize
\caption{Chord \RRTs{}, 15 runs per amplitude: maximum face slope, successes at 500 and 2000 iterations, first-solution iteration and share of candidate edges rejected.}
\label{tab:complete}
\setlength{\tabcolsep}{3.5pt}
\begin{tabular}{@{}lccccc@{}}
\toprule
Ampl. & Slope & 500 it. & 2000 it. & First sol. mean / worst & Rej. \\
\midrule
0   & $0^\circ$  & 15/15 & 15/15 & 72 / 114  & 0 \\
0.5 & $25^\circ$ & 15/15 & 15/15 & 73 / 110  & 0 \\
1   & $43^\circ$ & 15/15 & 15/15 & 73 / 116  & 0 \\
2   & $62^\circ$ & 15/15 & 15/15 & 71 / 120  & 0.1\% \\
4   & $75^\circ$ & 15/15 & 15/15 & 140 / 385 & 1.4\% \\
\bottomrule
\end{tabular}
\end{table}

\subsection{Choice of the sample count}
\label{sec:ksweep}

Both implementations may sample the chord at $k$ uniformly spaced points per step length, the chord for its tolerance test only, the surface primitive as extra starting points for its bisection; Table~\ref{tab:k} varies $k$ from 0 to 32 for both. For the surface primitive the reported cost is at or above the exact route length in all 360 runs, so $k$ does not affect consistency; route quality is flat or marginally best at small $k$, since a denser polyline follows the projection of the chord, which is longer than the geodesic between the nodes; time is flat within the measurement's scatter up to $k=4$ and grows beyond it, to 2.0--2.4 times the chord's at $k=32$; and the rejection share barely moves (1.2--1.5\% at amplitude 4). The primitive therefore uses none. For the chord, $k$ matters most at the first step and materially only on the roughest terrain: with no tolerance test ($k=0$) tunnelling chords are admitted and the routes are 0.85\% long, against 0.61\% at $k=1$ and 0.55\% at $k=3$; from $k=3$ to $k=32$ the route length moves by a further 0.07\% of $\cstar$ and the reported cost by 0.14 points, within 0.1\% of $\cstar$ and 0.2 points, at 1.3 times the cost, so the chord keeps $k=3$. The surface primitive's remaining constant, the bisection cap of six, was varied from 2 to 10 at amplitude 4: a shallow cap routes pairs that are not yet edge-adjacent through a shared vertex and lengthens the edges (routes at 1.0022 of $\cstar$ at depth 2 against 1.0016 at depth 6); from six to ten, route length, first solution and rejection share are unchanged to the precision reported (1.0016, 131 iterations, 1.2\%), and no pair lacked a shared vertex at the cap.

\begin{table}[t]
\centering\scriptsize
\caption{Both \textsc{Line} implementations against the sample count $k$: excess exact route length $\Delta\LS=(\LS-\cstar)/\cstar$ in units of $10^{-4}$, and time per iteration relative to the chord at $k=3$ (5 scenarios $\times$ 3 seeds). Better of the two in bold.}
\label{tab:k}
\setlength{\tabcolsep}{2.2pt}
\begin{tabular}{@{}llcccccccc@{}}
\toprule
Ampl. & & $k=0$ & 1 & 2 & 3 & 4 & 8 & 16 & 32 \\
\midrule
1 & surf.\ $\Delta\LS$ & \textbf{7} & \textbf{7} & \textbf{7} & \textbf{7} & \textbf{7} & \textbf{7} & \textbf{7} & \textbf{7} \\
  & surf.\ time & 1.23 & 1.23 & 1.26 & 1.31 & 1.31 & 1.44 & 1.62 & 1.95 \\
  & chord $\Delta\LS$ & 14 & 14 & 14 & 14 & 14 & 14 & 14 & 14 \\
  & chord time & \textbf{0.93} & \textbf{0.97} & \textbf{0.98} & \textbf{1.00} & \textbf{1.02} & \textbf{1.06} & \textbf{1.14} & \textbf{1.30} \\
\addlinespace
2 & surf.\ $\Delta\LS$ & \textbf{11} & \textbf{11} & \textbf{12} & \textbf{11} & \textbf{11} & \textbf{12} & \textbf{12} & \textbf{12} \\
  & surf.\ time & 1.39 & 1.40 & 1.43 & 1.53 & 1.55 & 1.62 & 1.88 & 2.31 \\
  & chord $\Delta\LS$ & 35 & 33 & 33 & 33 & 33 & 33 & 32 & 32 \\
  & chord time & \textbf{0.94} & \textbf{0.96} & \textbf{0.98} & \textbf{1.00} & \textbf{1.02} & \textbf{1.06} & \textbf{1.13} & \textbf{1.30} \\
\addlinespace
4 & surf.\ $\Delta\LS$ & \textbf{16} & \textbf{16} & \textbf{16} & \textbf{16} & \textbf{16} & \textbf{17} & \textbf{17} & \textbf{18} \\
  & surf.\ time & 1.52 & 1.45 & 1.43 & 1.40 & 1.47 & 1.55 & 1.85 & 2.38 \\
  & chord $\Delta\LS$ & 85 & 61 & 61 & 55 & 55 & 49 & 48 & 47 \\
  & chord time & \textbf{0.93} & \textbf{0.96} & \textbf{0.98} & \textbf{1.00} & \textbf{1.02} & \textbf{1.06} & \textbf{1.13} & \textbf{1.29} \\
\bottomrule
\end{tabular}
\end{table}

\subsection{Graph references}

\Astar{} on mesh edges is 7--10\% long everywhere, the familiar graph-metric penalty, and \Thetas{} with line-of-sight chords between vertices 2--6\% long on the curved terrains (Table~\ref{tab:baselines}); the \RRTs{} routes are on average 0.1--0.5\% long with chord edges and 0.07--0.16\% with surface edges.

\begin{table}[t]
\centering\footnotesize
\caption{Graph planners over the exact geodesic (mean of five scenarios).}
\label{tab:baselines}
\begin{tabular}{@{}lcc@{}}
\toprule
Ampl. & \Astar{} length$/\cstar$ & \Thetas{} $\LS/\cstar$ \\
\midrule
0   & 1.100 & 1.000 \\
0.5 & 1.088 & 1.019 \\
1   & 1.070 & 1.024 \\
2   & 1.085 & 1.039 \\
4   & 1.087 & 1.059 \\
\bottomrule
\end{tabular}
\end{table}

\section{Discussion and Limitations}
\label{sec:discussion}

\subsubsection*{Implications} The chord variant's routes, on average 0.1--0.5\% above the geodesic, are an order of magnitude better than \Astar{} on mesh edges; its reported cost is wrong by up to 2.7\% on average, in the direction no feasible path can have. On meshes the reported cost should be the surface length of the route and the \textsc{Line} primitive consistent in the sense of Section~\ref{sec:analysis}. The consistent primitive costs 1.2 to 1.5 times the chord's time per iteration, converges at the same rate in iterations and finds its first solution no later than the chord on average (131 against 140 iterations on the roughest terrain). Graph planners remain two to three orders of magnitude faster for a single query on meshes of this size; \RRTs{} remains worth having on a mesh for its anytime behaviour, because its start and goal need not be mesh vertices, and because it extends, like other sampling-based planners, to differential and kinodynamic constraints.

\subsubsection*{Limitations} All terrains are heightfields without holes or overhangs. On general meshes the Euclidean nearest neighbour can pick a node on an adjacent fold and closest-point projection can be ill-defined where a chord is far from the surface; the tolerance test rejects such edges, which keeps the result valid but wastes samples. Neither probabilistic completeness nor asymptotic optimality on the surface is proven for either variant: completeness rests only on the 75 of 75 successful runs, and optimality is argued from comparing geodesic disks with Euclidean balls under a slope bound and from the convergence of the surface-polyline cost to the geodesic as edges shrink. The meshes are small (529 vertices) and from one terrain family; three seeds per scenario are enough for the convergence means but too few for tail statistics.

\subsubsection*{Future work} A proof of asymptotic optimality for the surface primitive on heightfields; a variant that steers by exact geodesic tracing on the polyhedral surface, so that every parent edge is a geodesic segment; and experiments on scanned terrains and closed objects.

\section{Conclusion}
\label{sec:conclusion}

A surface has no volume in $\mathbb{R}^3$, so \RRTs{} as Karaman and Frazzoli define it cannot run on a terrain mesh. Mesh \RRTs{}, the reformulation on the surface that an implementation could make, solved every run on our heightfields, but with the naive chord edges it minimises the chord-polyline cost $\Jc$, which is strictly below the length of every feasible path; on the terrains tested the mean reported cost ends up as much as 2.7\% short of the true optimum, the shortfall grows with roughness, shrinks only linearly with the step, and cannot be removed by tightening the tolerance without disconnecting the graph, while the routes themselves are on average 0.1--0.5\% suboptimal. When each edge is instead a path on the surface through its projected points, charged its own length, the bias disappears, since the reported cost is the exact length of the route returned, which exceeds the geodesic by 0.07--0.16\% on average on every terrain tested, without any other modification to the algorithm.

\balance
\bibliographystyle{IEEEtran}
\bibliography{refs}

\end{document}